\documentclass{ecai}
\usepackage{times}
\usepackage{graphicx}
\usepackage{latexsym}

\usepackage{times}
\usepackage{helvet}
\usepackage{amsmath}
\usepackage{courier}
\usepackage{graphicx}

\usepackage{hyperref}      
\usepackage{url}           
\usepackage{booktabs}      
\usepackage{amsfonts}       
\usepackage{nicefrac}      
\usepackage{microtype}   
\usepackage{graphicx}
\usepackage{bm}
\usepackage{booktabs}

\begin{document}

\title{Knowledge-Enhanced Attentive Learning for Answer\\ Selection in Community Question Answering Systems}

\author{Fengshi Jing\institute{School of Data Science, City University of Hong Kong, Hong Kong SAR.} \and Qingpeng Zhang\institute{School of Data Science, City University of Hong Kong, Hong Kong SAR. Email: qingpeng.zhang@cityu.edu.hk. *Corresponding author} }

\maketitle
\bibliographystyle{ecai}

\begin{abstract}
  In the community question answering (CQA) system, the answer selection task aims to identify the best answer for a specific question, and thus is playing a key role in enhancing the service quality through recommending appropriate answers for new questions. Recent advances in CQA answer selection focus on enhancing the performance by incorporating the community information, particularly the expertise (previous answers) and authority (position in the social network) of an answerer. However, existing approaches for incorporating such information are limited in (a) only considering either the expertise or the authority, but not both; (b) ignoring the domain knowledge to differentiate topics of previous answers; and (c) simply using the authority information to adjust the similarity score, instead of fully utilizing it in the process of measuring the similarity between segments of the question and the answer. We propose the Knowledge-enhanced Attentive Answer Selection (KAAS) model, which enhances the performance through (a) considering both the expertise and the authority of the answerer; (b) utilizing the human-labeled tags, the taxonomy of the tags, and the votes as the domain knowledge to infer the expertise of the answer; (c) using matrix decomposition of the social network (formed by following-relationship) to infer the authority of the answerer and incorporating such information in the process of evaluating the similarity between segments. Besides, for vertical community, we incorporate an external knowledge graph to capture more professional information for vertical CQA systems. Then we adopt the attention mechanism to integrate the analysis of the text of questions and answers and the aforementioned community information. Experiments with both vertical and general CQA sites demonstrate the superior performance of the proposed KAAS model.
\end{abstract}

\section{INTRODUCTION}
Community question answering (CQA) systems can utilize the expertise of the community to provide timely and personalized service to Web users, and thus has merged as a key information acquisition platform for both general (e.g. Quora\footnote{https://www.quora.com/} and Zhihu\footnote{http://www.zhihu.com/}) and specific/vertical topics (e.g. HealthTap\footnote{https://www.healthtap.com/} and Stockoverflow\footnote{https://stackoverflow.com/}) [1, 2]. Since existing CQA sites have collected rich data of question answering pairs, it is possible to recommend an existing answer to a newly posted question. This is particularly important for “vertical” community for professional knowledge exchange and acquisition. For example, clinical doctors (those who write answers on HealthTap) cannot ensure their availability for 24 hours. If a patient posts a question in the middle of night, it is unrealistic to expect a doctor to answer it immediately. On the other hand, the question may have already been answered previously. Hence, recommending appropriate answers from the answer pool for new questions can provide needed information to the patient in a timely manner. In addition, we usually have more questions, many of which overlapping each other, than answers. These issues cause the question starvation problem, which requires effective answer selection models to enhance service quality through capitalizing the accumulated question/answer pool [3].

The answer selection [4] in CQA involves knowledge management and machine learning techniques with the primary focus on natural language processing (NLP) [5] and knowledge graphs (e.g. question classification [6] and measuring semantic similarity [7, 38]), because such online communities usually do not reveal the identity and detailed demographic information of users [8, 9, 10, 11, 12]. Typically, a long short-term memory (LSTM) framework is employed to learn the text representations and extract features [13, 14, 15, 16]. Attention mechanism has emerged as a common framework for such task due to its capability to capture the interrelations between different segments of the question and the answer [17, 18, 19]. To further enhance the performance, recent advances in CQA answer selection [2, 20, 21, 22, 23, 24] go beyond pure NLP by incorporating the community information, particularly the expertise (previous answers) and authority (position in the social network) of an answerer. Existing methods mainly adopt a two-phase approach, in which certain statistics are calculated and then imported into the downstream answer selection task. For instance, the count of an answerer's followers indicates his or her authority; the count of votes/agrees/thanks received by an answer indicates the quality of his or her previous answers [2, 23]. More recent studies utilize the text and tags of an answerer's previous answers to infer the domain expertise of the answerer [20, 21, 22, 24] . Despite being effective, existing methods for incorporating community information are limited:

(a) They only consider either the expertise or the authority, but not both. In practice, both types of information may contribute to predicting the quality and relevance of an answer.

(b) Domain knowledge is not fully utilized to differentiate topics of previous answers. To be specific, previous studies [21, 22] depict the expertise of an answerer through analyzing the textual information of all the answerer's previous answers. This approach is appropriate for vertical CQA communities since the answerers (e.g. a clinical doctor) are not likely to answer questions that are irrelevant to their expertise. However, this approach is limited for general CQA communities, in which users often answer various questions. If the full text of an answerer's previous answers is used, we may include irrelevant expertise information. For example, a user's previous answer about Chinese history does not provide any information for a question about a machine learning algorithm. In addition, using the full text of all previous answers requires huge computational resources.

(c) The expertise and authority information (e.g. the count of followers and the expertise of the followers) is only used to adjust the final similarity score (e.g. [22, 24]), instead of being fully utilized in the process of measuring the similarity between segments of the question and the answer.

More specifically, existing approaches assume that an answerer's authority can be represented by the linear combination of followers' specialties. The similarity between the question and an answerer is be derived by a model first and then adjusted by the answerer's authority. This is problematic because different specialties are not entirely independent to each other; there could be interrelations among two or more specialties. In addition, such expertise and authority information could have been fully utilized in the NLP process that evaluates the similarities among segments of questions and answerers.

\begin{figure}
 \centering
  \includegraphics[scale=0.345]{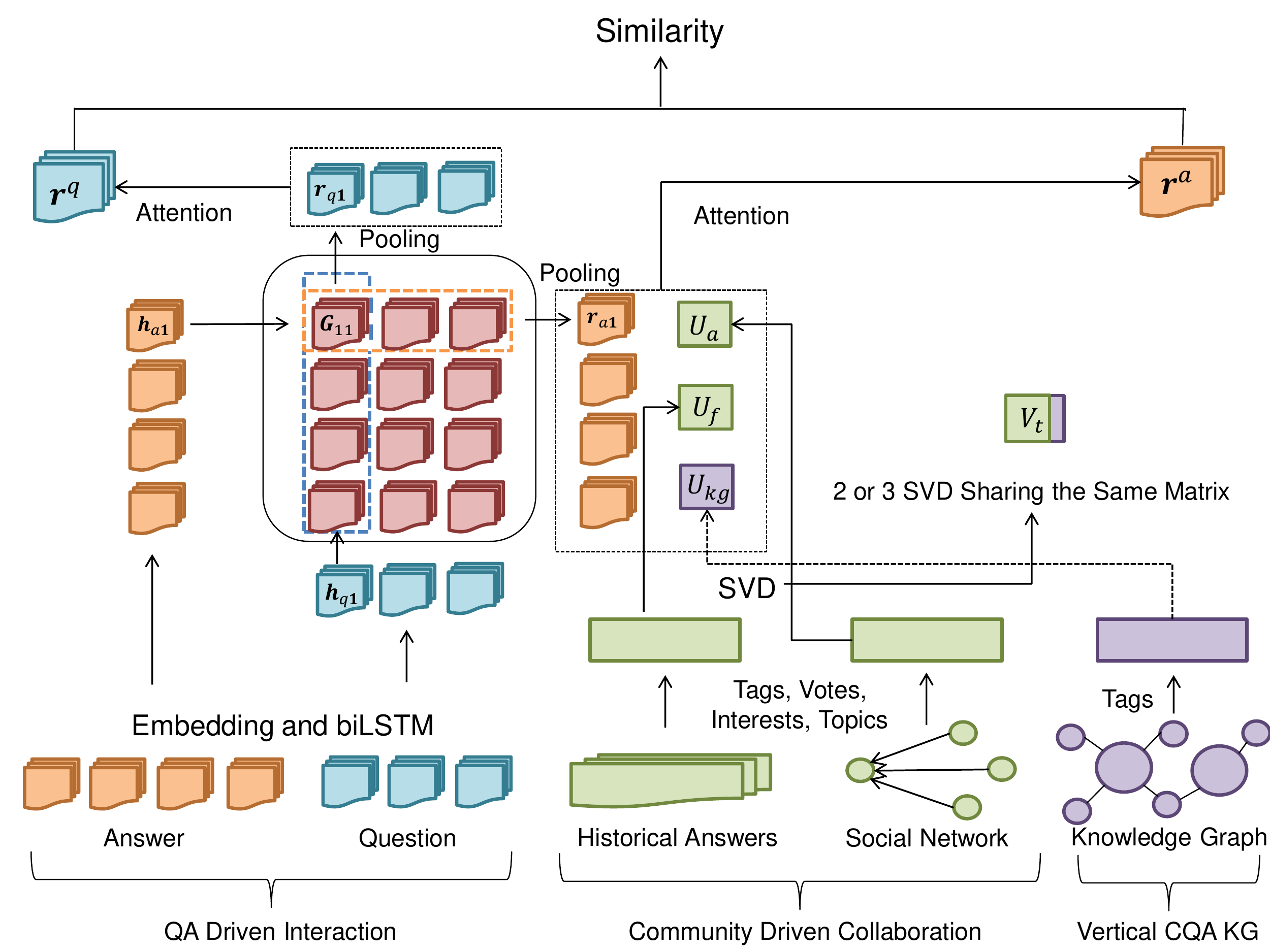}
  \caption{Framework of the KAAS model.}
\end{figure}

To fill the aforementioned gaps, we have proposed the Knowledge-enhanced Attentive Answer Selection (KAAS) model (framework is shown by \textbf{Figure 1}). First, we introduce an expertise matrix and an authority matrix to capture the expertise information from historical answers and the authority information from the answerer's followers in the social networks, respectively. The two matrices share the same (human-labeled) tag dimension, which represents the predefined topic/specialty structure. To address the sparsity problem caused by the large number of tags, we utilize the taxonomy of tags to group tags that are semantically similar to each other. Second, we extract the latent feature matrices for expertise and authority through decomposing the two corresponding matrices. Third, we adopt an attention mechanism to examine the similarity among segments of questions and answers. The answer's attention representation is adjusted by the extracted expertise and authority features of the answerer.  Eventually, the similarity between the question and a candidate answer is generated by taking the inner product of the attentive representation of the question and the adjusted attentive representation of the answer. Experiments with both vertical and general CQA sites demonstrate the superior performance of the proposed KAAS model. For vertical community (i.e., HeathTap, which is a medical area CQA site), we introduce an  external knowledge graph (i.e., health knowledge graph) and embed this knowledge graph before attentive learning.

There are three main contributions of our paper. First, KAAS incorporates both the expertise and the authority of the answerer to enhance the performance. Second, we utilize the human-labeled tags and votes as the domain knowledge to infer the expertise of the answer. Third, we propose a matrix decomposition-based method to infer the authority of the answerer and incorporate such information in the process of evaluating the similarity between segments. 

\section{RELATED WORK}

\subsection{The general neural network framework for answer selection}

To capture the sequential contextual features in free text, LSTM [25], particularly bidirectional LSTM (biLSTM) [26], has been the basic modeling framework for answer selection [27]. \textbf{Figure 2a} presents this general framework. First, an embedding method (e.g. word2vec [28]) is used to encode the text. Second, we use biLSTM to generate the question feature matrix $\textbf{Q}$ and answer feature matrix $\textbf{A}$, respectively. Third, column-wise max pooling (or other pooling methods) is used to transform the feature matrices into the representing vectors of the question and answer. Last, the similarity score is obtained by calculating the cosine similarity between the representing vectors.

There are a number of variants that improve over the base biLSTM model, including replacing biLSTM with convolutional neural network (CNN) as the sentence model [29], using both CNN and biLSTM to jointly learn feature matrices [30], and using multiple biLSTM components to learn the similarity from the feature matrices/vectors [31].

\begin{figure}
 \centering
  \includegraphics[scale=0.37]{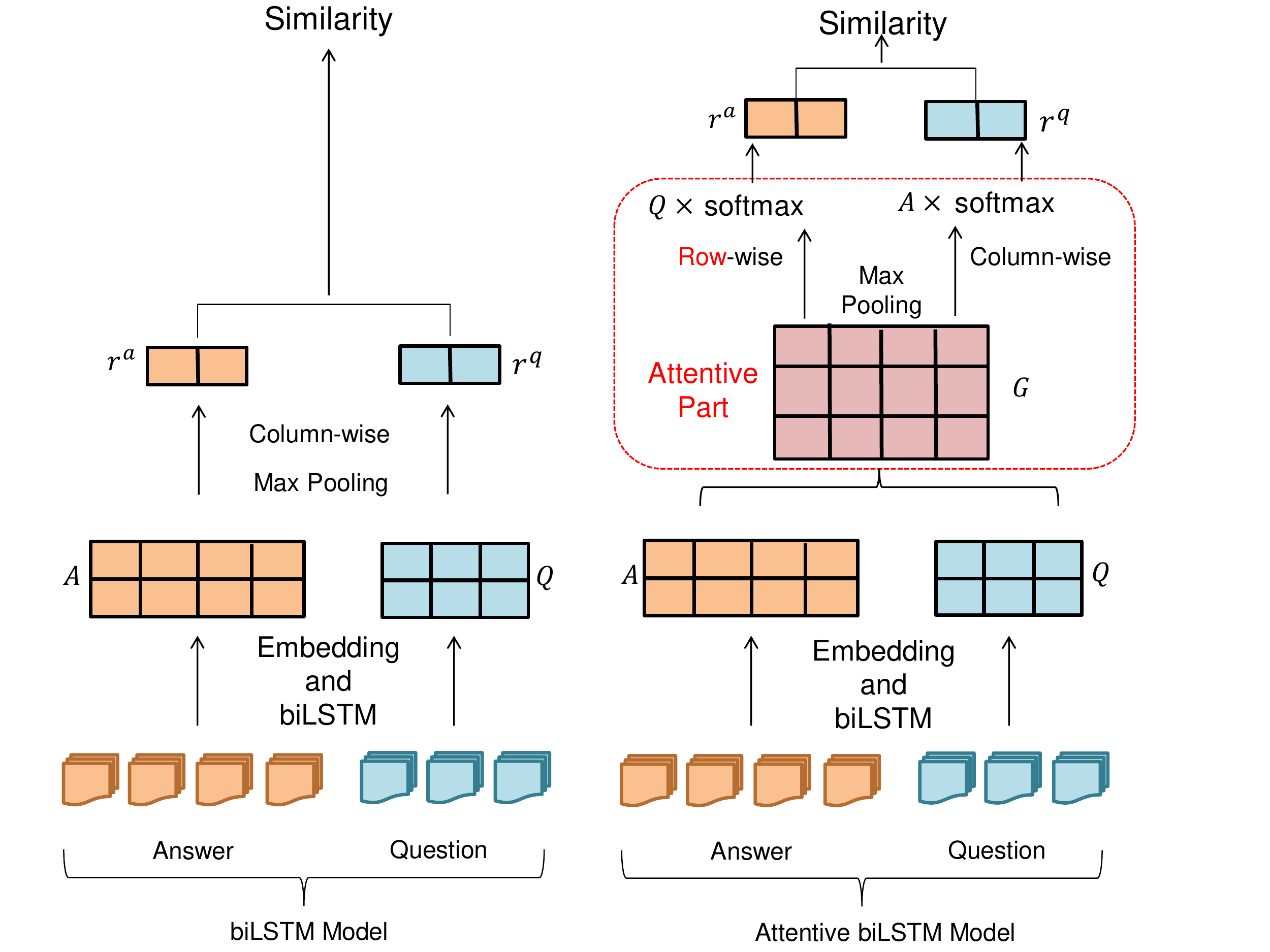}
  \caption{The general neural network framework for answer selection.}
\end{figure}

\subsection{The attentive framework for answer selection}

Attentive pooling is a method to enable the pooling layer to be aware of the input pair [18]. With an attention mechanism, the information of input items directly influence the calculation of each other's representations, and thus enhance the capability in evaluating the similarity between two inputs [31]. The attentive pooling method (shown in \textbf{Figure 2b}) has been recently adopted as a standard for answer selection task, in which the two inputs can naturally represent the question matrix and answer matrix (from the biLSTM component) [32, 33]. A recent study further extends the attention matrix to 3rd-order tensor to consider the relationships among segments within questions or answers [23].

\subsection{Recent advances in incorporating community information}

Recent advances [2, 20, 21, 22, 23, 24] in CQA answer selection go beyond pure text mining and incorporate the community information. Particularly, we can evaluate the answerer's expertise through analyzing his or her previous answers, and estimate the authority of an answerer through examining his or her topological position in the community. Zhao et. al (2017) propose the Asymmetric Multi-Faceted Ranking Network Learning (AMRNL) model [20] that uses the count of an answerer's followers to indicate the authority, and adjust the similarity score by introducing the question-authority matching score [24]. Lei et. al (2018) borrow the idea of residual networks and propose the Multi-View Fusion Neural Network (MVFNN) model to take topics of the question into consideration [17]. Wen et. al (2019) propose the Hybrid Attentive with Deep Users (UIA-LSTM) model that combines the text of the candidate answer and the text of the corresponding answerer's previous answers for the following attentive pooling procedure [21, 22]. This approach is effective for vertical CQA communities, where users often do not answer questions out of their domain expertise. However, it might introduce noise for the answer selection task in general CQA communities because users may answer questions in quite different domains.

\section{MODEL}

\subsection{Text representation}
First, we perform the word embedding of the original text. Word2vec [28] is used to train the word vector. Note that we may use other word embedding methods. Next, we follow [27] to use the biLSTM to represent the question and the answer. More specifically, we represent a given sentence as $X=(x_{1},x_{2},\ldots,x_{n})$, in which $x_{t}$ is the $100$-dimension embedded vector for the word. The hidden vector $h_{t}$ at time step $t$ in the LSTM component is updated as follows:
\begin{eqnarray} 
&& i_{t}=\sigma(W_{i}x_{t}+U_{i}h_{t-1}+b_{i}),\\ 
&& f_{t}=\sigma(W_{f}x_{t}+U_{f}h_{t-1}+b_{f}),\\
&& o_{t}=\sigma(W_{o}x_{t}+U_{o}h_{t-1}+b_{o}),\\
&& C_{t}^{'}=\tanh(W_{c}x_{t}+U_{c}h_{t-1}+b_{c}),\\
&& C_{t}=i_{c}*C_{t}^{'}+f_{t}*C_{t-1}),\\
&& h_{t}=o_{c}*\tanh(C_{t}),
\end{eqnarray} 
where $\sigma$ is the sigmoid activation function, $i$ represents the input gate, $f$ represents the forget gate, $o$ represents the output gate, $C$ denotes the cell memory, and $W$, $U$, $b$ are network parameters. The formula (4) is the input transformation, and formula (5) updates of the cell state.

The standard LSTM only uses the information of the past. biLSTM, on the other hand, utilizes both the previous and future context by processing the sequence on two directions, and generates two independent sequences of LSTM output vectors.  Because biLSTM models the context information for each word, biLSTM-based representation is usually more accurate than LSTM in the answer selection task [18]. In our model, the biLSTM output at each time step is the concatenation of the two output vectors from both directions, i.e., $h_{t}=h_{t}^{f}||h_{t}^{r}$.

\subsection{Expertise representation based on previous answers}
Previous answers provided by an answerer can be used to represent the expertise of the answerer. Particularly, the CQA site usually has the \textit{tag} function to help users manually label the questions and answers into a predefined categorical system of the domain knowledge. Such rich information can help us model the answerers' expertise with a high resolution (as compared to only using the text in the answers). We adopt a collaborative filtering [34, 35] approach to capture the relationship between a tag and an answerer based on the relationships among all tags and previous answers. Because of the large number of tags, there exists the sparsity problem that we do not have sufficient data to learn the representation of each tag [36]. Therefore, we make use of the taxonomy of the tags to group semantically similar tags to a single higher-level tag. We illustrate the grouping procedure in \textbf{Figure 3} using the two datasets used in this study. 

\begin{figure}
 \centering
  \includegraphics[scale=0.3]{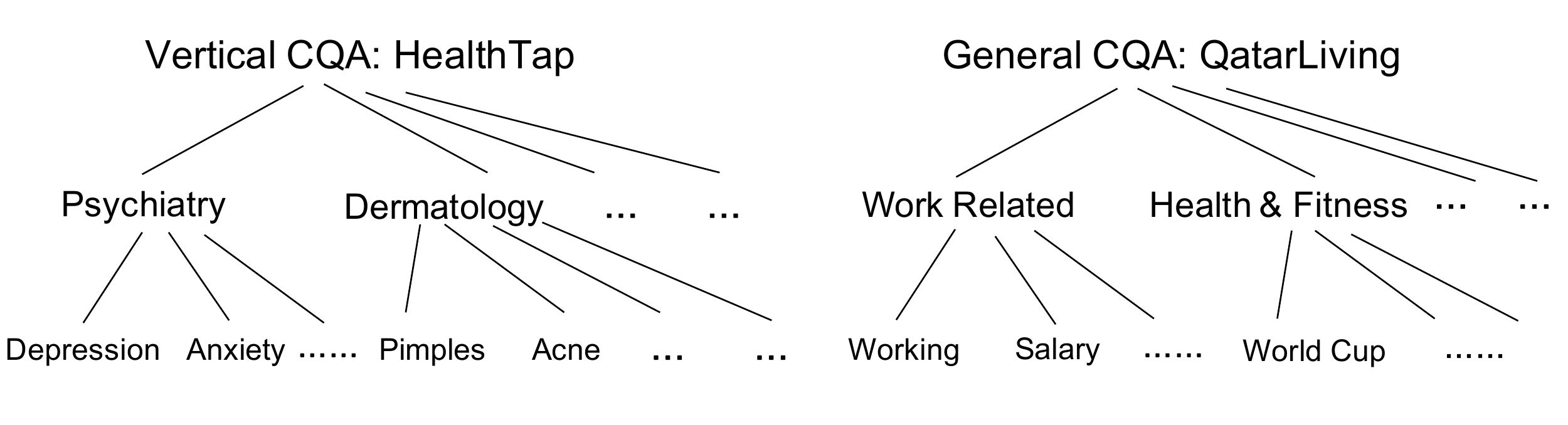}
  \caption{Grouping procedure using the two datasets. For both CQA communities, we group the lower-level tags to the immediate higher-level tags. For example, "depression" and "anxiety" are grouped to "psychiatry," and "working" and "salary" are grouped to "work related."}
\end{figure}

After the tag grouping process, we define a weight $H_{ij}^a$ to represent answerer $a$'s expertise in tag $j$ as expressed by answer $i$. $H_{ij}^a$ is measured by the product of the frequency of a certain tag in an answer and the vote measure for this answer as follows
\begin{eqnarray} 
&& H_{ij}^a=f_i(j)\cdot v(i),
\end{eqnarray} 
where $a$ denotes the answerer, $f_i(j)$ denotes the frequency of tag $j$ in the previous answer $i$, and $v(i)$ denotes the vote measure for answer $i$. For brevity, we omit the answerer identifier $a$ in the rest of the paper since we focus on modeling the answers and follower of a single answerer (no interactions between competing answerers). For the CQA community (HealthTap) that exhibits the actual count of up-votes, the vote measure is the count. For the CQA community (QatarLiving\footnote{https://www.qatarliving.com/forum}) that only provides a categorical evaluation of the quality of answer (i.e. "Good"/"Potentially Useful"/"Bad"), we set a numeric value for each category: “Good”=2, “Potentially Useful”=1, and “Bad”=0. For an answerer, we construct an answer-tag quality matrix $\textbf{H}\in \mathbb{R}^{A\times T}$. $A$ is the number of answers posted by the answerer, and $T$ is the total number of tags that are shared by all answers. Then, we employ the singular-value decomposition (SVD) [37] to decompose $\textbf{H}$ to obtain the feature matrices for the answerer's previous answers (expertise) and the tags as follows
\begin{eqnarray} 
&& \textbf{H}=U_{a}\Sigma_{a}V_{a},
\end{eqnarray} 
where $U_{a}$ represents the feature matrix for the answerer's previous answers, $\Sigma_{a}$ represents the scaling matrix, and $V_{a}$ represents the feature matrix for the tags.

\subsection{Authority representation based on social networks}
The topological position of an answerer in the social network formed by following relationship represent the authority of the answerer. In addition, the interests/specialty of an answerer's followers (expressed by tags) can be used to further enrich the representation of the answerer's authority. Existing method of inferring the answerer's authority is to have a linear combination of the followers' tags [34, 35]. However, this approach has a strong assumption that the tags are independent. Similarily, we employ the SVD [37] to capture the relationship between the an answerer and his or her followers (and their tags) based on the relationships among all tags and followers. Semantically similar tags are also grouped to a single higher-level tag (as shown in \textbf{Figure 3}).  

After the tag grouping process, we define a weight $S_{ij}$ to represent the answerer's authority in tag $j$ as inferred from the answerer's follower $i$. $S_{ij}$ is measured by the frequency of a certain tag of a follower as follows
\begin{eqnarray} 
&& S_{ij}=f_i(j),
\end{eqnarray} 
where $f_i(j)$ denotes the frequency of tag $j$ of the answerer's follower $i$. Note that this tag is labeled by the follower $i$. Before tag grouping, the frequency for each tag is either 1 or 0. After tag grouping, the frequency refers to the number of lower-level tags labeled by $i$. For example, if the follower is labeled with "depression" and "anxiety," both are then grouped to "psychiatry," the frequency of "psychiatry" for $i$ is 2. For the follower $i$, we construct a follower-tag quality matrix $\textbf{S}\in \mathbb{R}^{F\times T}$. $F$ is the number of the answerer's followers and $T$ is the total number of tags that are shared by all. Then, we employ the SVD to decompose $\textbf{S}$ to obtain the feature matrices for the answerer's social network and the tags as follows
\begin{eqnarray} 
&& \textbf{S}=U_{s}\Sigma_{s}V_{s},\\
&& V_{s}=V_{a},
\end{eqnarray} 
where $U_{s}$ represents the feature matrix for the answerer's social network (authority), $\Sigma_{s}$ represents the scaling matrix, and $V_{s}$ represents the feature matrix for the tags which is set equal to $V_{a}$.

As for the tag matrix, because the SVD is not unique as a small portion of top singular values can approximately represent the whole matrix, so we first approximate matrix $V_a$, and then set the $V_s$ and $V_kg$ equal. SVD is set based on numeric experiments.

\subsection{Knowledge graph representation}
In vertical community question answering system, there are much professional knowledge. To make full use of this kind of knowledge, we incorporate an external knowledge graph. For example, in this paper, for HealthTap site, we introduce a health knowledge graph (as shown in \textbf{Figure 4}) which is derived from Electronic Medical Records (EMR) [43] and it can include more professional medical knowledge. There are high quality knowledge bases linking diseases and symptoms, while diseases can be grouped to single higher-level tags similarly. Then we define a weight $KG_{ij}$ to represent the candidate answer's total weights in tag $j$ in terms of the answer's symptom concept $i$. $KG_{ij}$ is measured by the frequency of a certain tag of symtoms as follows
\begin{eqnarray} 
&& KG_{ij}=w_i(j),
\end{eqnarray} 
where $w_i(j)$ denotes the total weights of tag $j$ in terms of the answer's symptom concept $i$. For the symptom $i$, we construct a symptom-tag quality matrix $\textbf{KG}\in \mathbb{R}^{SC\times T}$. $SC$ is the number of the answer's symptom concepts and $T$ is the total number of tags that are shared by all. Then, we employ the SVD to decompose $\textbf{KG}$ to obtain the feature matrices for the answer's symptom concepts and the tags as follows
\begin{eqnarray} 
&& \textbf{KG}=U_{kg}\Sigma_{kg}V_{kg},\\
&& V_{s}=V_{a}=V_{kg},
\end{eqnarray} 
where $U_{kg}$ represents the feature matrix for the answer's symptom concepts (relationship knowledge graph), $\Sigma_{kg}$ represents the scaling matrix, and $V_{kg}$ represents the feature matrix for the tags which is set equal to $V_{a}$ and $V_{s}$.
\begin{figure}
 \centering
  \includegraphics[scale=0.75]{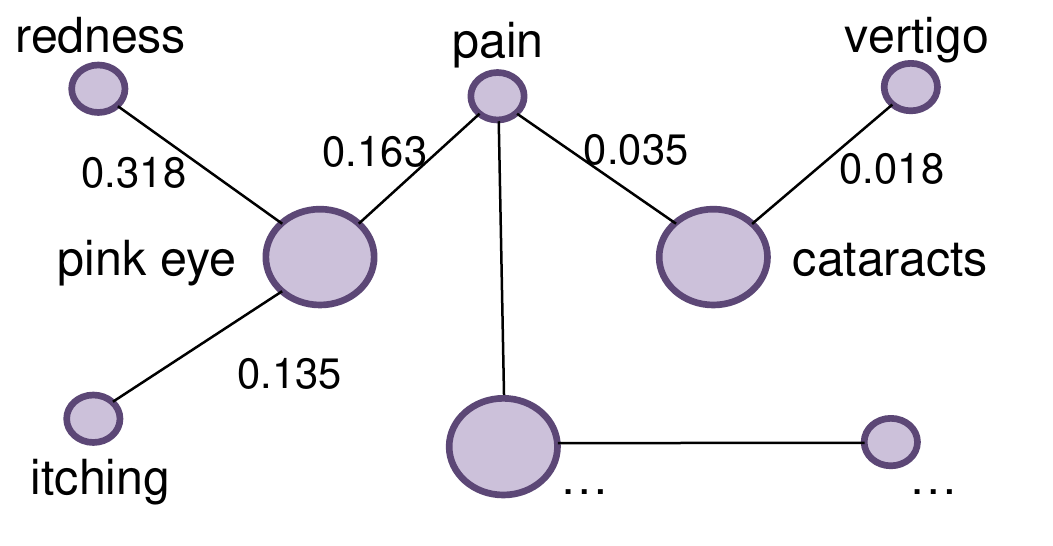}
  \caption{Health Knowledge Graph [43]. Different diseases (i.e., lower-level tags; e.g., pink eye, cataracts) have a relating graph to the symptom concepts (e.g., pain, redness, vertigo). From this knowledge graph, we can see that ophthalmic (a higher-level tag) has a sum relationship weight $0.198=0.035+0.163$ with pain (a symptom concept which might be mentioned in answers).}
\end{figure}

\subsection{Question-answer pair concatenation}
To capture the relationship between each segment in the question and each segment in the answer, we follow [23] to construct a concatenation tensor $\textbf{G}$ as follows
\begin{eqnarray} 
&& G_{ij}=\sigma(W_{1}*[h_{q(i)},h_{a(j)}]+b_{1}),
\end{eqnarray} 
where $i=1,2,\ldots,M;j=1,2,\ldots,L$ and $\sigma$ is the sigmoid activation function, $W_{1}$ is the transformation matrix, $h_{q(i)}$ is the $i$-th hidden state of biLSTM question representations, $h_{a(j)}$ is the $j$-th hidden state of biLSTM answer representations, and $b_{1}$ is the bias vector. $M$ is the length of the question and $L$ is the length of the answer. 

Following the standard attentive pooling scheme [18], we use the row-wise pooling to obtain an interaction matrix $\textbf{r}_q$, which captures the relationship between each segment in question with all segments in answer, and use the column-wise pooling to obtain an interaction matrix $\textbf{r}_a$, which captures the relationship between each segment in answer and all segments in the question
\begin{eqnarray} 
&& r_{q(i)}=\max_{1<l<L}G_{il},\\
&& r_{a(j)}=\max_{1<m<M}G_{mj}.
\end{eqnarray} 

\subsection{Attention calculation and similarity}
For each candidate answer, we model each answerer's expertise and authority using $U_{a}$ and $U_{s}$, respectively. Then we combine them with $r_{a(j)}$ as: $y_{a(j)}=\tanh(W_{2}r_{a(j)}+U_{a}W_{3}+U_{s}W_{4}+U_{kg}W_{5}+b_{2})$, where $y_{a(j)}$ is the unnormalized attention of the $j$-th segment in the answer, $W_{2}$, $W_{3}$ and $W_{4}$ are transformation matrices, and $b_{2}$ is the bias vector. Then, the formal attention $\alpha_{a(j)}$ of each segment in a answer is calculated by
\begin{eqnarray} 
&& \alpha_{a(j)}=\mathrm{softmax}(e^{y_{a(j)}}).
\end{eqnarray} 
The final representation of the answer is calculated as the weighted summation of the interactions
\begin{eqnarray} 
&& r^{a}=\sum_{j=1}^{L}\alpha_{a(j)}r_{a(j)},\\
\end{eqnarray} 
The final representation of the question $r^{q}$ is directly derived from the attentive pooling as follows
\begin{eqnarray} 
&& \alpha_{q(i)}=\mathrm{softmax}(e^{r_{q(i)}}),\\
&& r^{q}=\sum_{i=1}^{M}\alpha_{q(i)}r_{q(i)},
\end{eqnarray} 
where $\alpha_{q(i)}$ is the attention of each segment in the question.

Eventually, the similarity score of the question and the answer is defined as follows
\begin{eqnarray} 
&& s(q,a)=\tanh[(r^{q})^{T}U_{r}r^{a}],
\end{eqnarray} 
where $U_{r}$ is a parameter matrix.

For each question, we identify two question-answer pairs, $(q,a^{+})$ and $(q,a^{-})$, with $a^{+}$ denotes the best answer and $a^{-}$ denotes the worse answer. We formulate a WARP loss function to optimize the top of the answer ranking [38]. 
\begin{eqnarray} 
&& L=\max(0,m+s(q,a^{-})-s(q,a^{+})).
\end{eqnarray} 

\section{EXPERIMENTS}
\subsection{Data and model setup}
HealthTap is a medical (vertical) CQA site with over 111,000 certified clinical doctors who have answered over 6.5 billion questions by 2018. We adopt the benchmark HealthTap data provided by [2]. We sample all the 4,781 questions with at least two answers and the vote difference between the best and worse answers is at least two. 500 questions (around 10\%) are randomly selected as the test set. For consistency, we follow a commonly adopted approach to make the count of answers for each question to be 20. If a question has more than 20 answers, we select the top 20. If a question has fewer than 20 answers, we randomly choose answers [39] from the answer pool to make it 20. We perform five-fold cross-validation to avoid the overfitting issue. QatarLiving is a general CQA site about various topics of living in Qatar [40]. It is a commonly used benchmark dataset for answer selection. There are 5,450 questions and 41,908 answers. Following the benchmark rules [40], 10.48\% are selected as the test set.

The KAAS model is implemented with the Tensorflow (>=1.8) framework in Python 2.7. Word2vec is used for training the word vector with the vector dimension of 100. The maximum lengths for the question and the answer are 40 and 80, respectively. The optimization method is Stochastic Gradient Descent (SGD) with the batch size of 256 and the learning rate of 0.01. The value of margin $m$ is set as 0.05. The hidden size of the biLSTM is 128. We use the Top 1 precision (P@1), Top 2 precision (P@2), MAP, Accuracy and F1-score to evaluate the performance of KAAS. P@K stands for the proportion of the selected answers in the top K that are true. Usually, users will pay much attention to top 1 answer, sometimes to top 2, and few times to answers behind the second. As a result, we consider about P@1 and P@2. For HT dataset, the ground truth is the number of votes, which ranks the answers from the most-voted to the least-voted. So we use P@1 and P@2 to evaluate the performance of our model for HT dataset. For QL dataset, the ground truth is the human-labeled group (i.e., “Good”, “Potentially Useful” and “Bad”), which provides with answers' quality classification and that's why MAP, Accuracy and F1-score are used for QL dataset. We run the experiments on a Linux server with two E5-2630 v4 2.2GHz CPU and 64GB RAM. The source code is released\footnote{https://sites.google.com/view/jingfengshi/home/blog/code}.

\subsection{Baseline models}
We compare the performance of \textbf{KAAS} with that of following state-of-the-art models. \textbf{PLANE} is a non-neural network method based on statistical NLP feature extractions. It has an offline learning component and an online search component [2]. \textbf{LSTM} is the basic biLSTM model without attentive component [27]. \textbf{AP-LSTM} has a similar biLSTM architecture with the attentive pooling component [18]. \textbf{AI-CNN} takes the interaction of sentence pair into consideration, resulting in a 3D tensor to capture the relationship among the segments [23]. \textbf{AI-CNN-F} computes the similarity through adding additional community information (received thanks and agrees) [23]. \textbf{MVFNN} models answer selection task with a multi-view fusion neural network based on the idea of residual networks [17]. \textbf{AMRNL} uses a linear combination of followers' tags to represent the authority information, and adjusts the final matching score [20].

\subsection{Results}
\begin{table}
\centering
\caption{Performance on HealthTap (HT) dataset.}
\begin{tabular}{@{}ccc@{}}
\toprule
Model\textbackslash{}Evaluation & P@1             & P@2           \\ \midrule
PLANE                           & 33.2\%          & 52.6\%          \\
LSTM                            & 37.7\%          & 59.5\%          \\
AP-LSTM                         & 39.5\%          & 62.4\%          \\
AI-CNN                          & 39.8\%          & 62.9\%          \\
AI-CNN-F                        & 40.1\%          & 63.7\%          \\
MVFNN                           & 40.1\%          & 63.5\%          \\
AMRNL                           & 39.4\%          & 62.9\%          \\
KAAS                            & \textbf{40.4\%} & \textbf{64.8\%}\\ \bottomrule
\end{tabular}
\end{table}

\begin{table}
\centering
\caption{Performance on QatarLiving (QL) dataset.}
\begin{tabular}{@{}cccc@{}}
\toprule
Model\textbackslash{}Evaluation & MAP             & Accuracy        & F1              \\ \midrule
PLANE                           & 69.7\%          & 66.5\%          & 60.0\%         \\
LSTM                            & 75.3\%          & 74.1\%          & 70.9\%          \\
AP-LSTM                         & 77.1\%          & 75.5\%          & 71.7\%          \\
AI-CNN                          & 79.2\%          & 76.3\%          & 72.8\%          \\
AI-CNN-F                        & 80.1\%          & 76.9\%          & 73.0\%          \\
MVFNN                           & 80.0\%          & 77.3\%          & 73.3\%          \\
AMRNL                           & 70.4\%          & 68.5\%          & 62.7\%          \\
KAAS                            & \textbf{80.7\%} & \textbf{78.6\%} & \textbf{74.4\%}\\ \bottomrule
\end{tabular}
\end{table}

Table 1 and 2 present the performance on both HealthTap and QatarLiving datasets. In general, the proposed KAAS model consistently outperforms state-of-the-art baseline models. More specifically, for the HealthTap dataset, we use P@1 and P@2 to measure the accuracy. P@K is the frequency of successfully predicting the best $K$ answer. As the only non-neural network model, PLANE has the lowest accuracy, but it has the advantage in computational efficiency. From LSTM to AP-LSTM and to AI-CNN/AI-CNN-F, the performance improves with additional attentive pooling framework and community information. MVFNN's performance is similar to AI-CNN-F because it also utilizes the simple community information. AMRNL, on the other hand, only leads to similar performance as the AP-LSTM, indicating the linear combination of tags is less effective than expected, probably due to the sparsity problem in the tag distributions. 

Because the QatarLiving dataset provides a categorical evaluation (instead of vote count) of the answer, we adopted MAP, Accuracy, and F1-score to evaluate the performance. We have similar finding: the attentive pooling framework and the inclusion of community information can improve the performance. The proposed KAAS model performs the best consistently.

\begin{figure}
 \centering
  \includegraphics[scale=0.36]{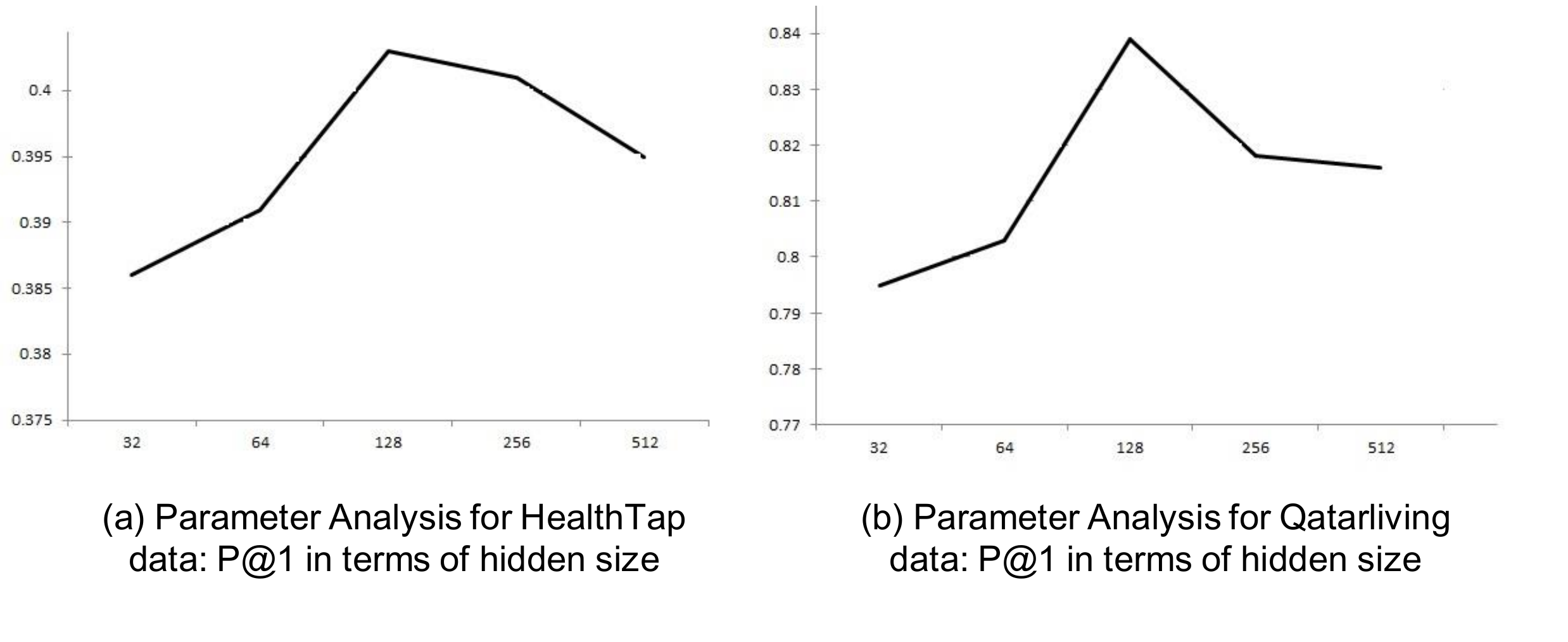}
  \caption{Parameter analysis for two datasets}
\end{figure}

We further analyze the sensitivity of the KAAS model in terms of the size of the biLSTM hidden layer. As shown in \textbf{Table 3} and \textbf{Figure 5}, we observe that the size of the hidden layer influences the performance of the model. A trade-off lies between the model complexity and the performance. In particular, when the hidden layer size is small, we can improve the performance by increasing the size of the hidden layer. However, when the hidden layer size is large than a change point, the performance declines, which could be due to the overfitting issue and the lack of sufficient data to fit the additional parameters. The size of 128 in the previous experiments is set based on this sensitivity analysis. Finally we conduct ablation studies as shown in \textbf{Table 4}.  We can find that among three parts (i.e., authority, expertise and knowledge graph), expertise information is the most significant. In the future, we plan to optimize the matrix decomposition part to further enhance the full model's performance.

\begin{table}
\caption{The performance with different sizes of the hidden layer.}
\centering
\begin{tabular}{@{}cccccc@{}}
\toprule
Hidden Size & 32     & 64     & 128             & 256    & 512    \\ \midrule
P@1 HT      & 38.8\% & 39.1\% & \textbf{40.4\%} & 40.2\% & 39.4\% \\
P@1 QL      & 79.5\% & 80.3\% & \textbf{83.9\%} & 81.8\% & 81.6\% \\ \bottomrule
\end{tabular}
\end{table}

\begin{table}
\caption{Results of ablation studies in HT Dataset.}
\centering
\begin{tabular}{@{}cc@{}}
\toprule
Model                             & P@1     \\ \midrule
KAAS None                         & 39.60\% \\
KAAS Expertise                    & 40.23\% \\
KAAS Authority                    & 39.95\% \\
KAAS Knowledge Graph              & 39.73\% \\
KAAS Expertise \& Authority       & 40.32\% \\
KAAS Authority \& Knowledge Graph & 40.07\% \\
KAAS Knowledge Graph \& Expertise & 40.11\% \\
KAAS Full                         & 40.38\%\\ \bottomrule
\end{tabular}
\end{table}

\section{CONCLUSIONS}
In this paper, we propose the KAAS model for the CQA answer selection task. KAAS is based on a biLSTM neural network with attentive pooling mechanism. It incorporates both the expertise and authority information learned from the answerer's previous answers, the tags of the answers, the followers of the answerer, and the tags of the followers. For vertical community, an external knowledge graph is also utilized which can capture semantic information between questions and answers [41, 42, 43]. In the end, experiments with both general and vertical CQA datasets show that the KAAS model outperforms state-of-the-art answer selection models.

The novelty of our model comes from the incorporation of community information and domain knowledge. We combine the existing techniques with an efficient modeling framework. The model presents a generic framework that incorporates the community information as well as external knowledge, which does not necessarily have to be in the same format in different datasets. We can easily modify the SVD and attention components to incorporate different types of community information that can be extracted from other datasets or other types of side information. For CQA sites where no authors' information is available, we can always incorporate external knowledge graphs into it or we may crawler information from website. In hence, our proposed model is both general and novel, and we can rather easily apply it to other recommendation problems in CQA .

In conclusion, this paper sheds light on the efficacy of the community information in inferring the expertise and authority of answerers, and could inform future research to better mine and utilize the domain knowledge hidden in the CQA community. One possible future direction is to optimize the decomposition process. Given that we will have to consider its influence on downstream learning task, the optimization is difficult but a valuable try. Another opportunity is to generate knowledge graph from the CQA community itself, which might be more suitable for revealing semantic information between the community questions and answers, and thus is a rewarding while very challenging future work. 

\section*{REFERENCES}
\medskip
\small

[1] Yuan, S., Zhang, Y., Tang, J., Hall, W.\ \& Cabotà, J. B.\ (2019). Expert finding in community question answering: a review. {\it Artificial Intelligence Review}, in press.

[2] Nie, L., Wei, X., Zhang, D., Wang, X., Gao, Z.\ \& Yang, Y.\ (2017). Data-driven answer selection in community QA systems. {\it IEEE transactions on knowledge and data engineering} 29(6), 1186-1198.

[3] Nie, L., Wang, M., Zhang, L., Yan, S., Zhang, B.\ \& Chua, T. S.\ (2015). Disease inference from health-related questions via sparse deep learning. {\it IEEE Transactions on knowledge and Data Engineering}, 27(8), 2107-2119.

[4] Sun, R., Cui, H., Li, K., Kan, M. Y.\ \& Chua, T. S.\ (2005). Dependency relation matching for answer selection. In {\it Proceedings of the 28th annual international ACM SIGIR conference on Research and development in information retrieval} (pp. 651-652). ACM.

[5] Zhang, D.\ \& Lee, W. S.\ (2003). Question classification using support vector machines. In {\it Proceedings of the 26th annual international ACM SIGIR conference on Research and development in informaion retrieval} (pp. 26-32). ACM.

[6] Manning, C. D., Manning, C. D.\ \& Schütze, H.\ (1999). {\it Foundations of statistical natural language processing}. Cambridge, MA: MIT Press.

[7] Moschitti, A., Quarteroni, S., Basili, R.\ \& Manandhar, S.\ (2007). Exploiting syntactic and shallow semantic kernels for question answer classification. In {\it Proceedings of the 45th annual meeting of the association of computational linguistics} (pp. 776-783). ACL.

[8] Heilman, M.\ \& Smith, N. A. (2010).\ Tree edit models for recognizing textual entailments, paraphrases, and answers to questions. In {\it Human Language Technologies: The 2010 Annual Conference of the North American Chapter of the Association for Computational Linguistics} (pp. 1011-1019). ACL.

[9] Xue, X., Jeon, J.\ \& Croft, W. B. (2008).\ Retrieval models for question and answer archives. In {\it Proceedings of the 31st annual international ACM SIGIR conference on Research and development in information retrieval} (pp. 475-482). ACM.

[10] Rao, J., He, H.\ \& Lin, J. (2017).\ Experiments with convolutional neural network models for answer selection. In {\it  Proceedings of the 40th International ACM SIGIR Conference on Research and Development in Information Retrieval} (pp. 1217-1220). ACM.

[11] Qiu, X.\ \& Huang, X. (2015).\ Convolutional neural tensor network architecture for community-based question answering. In {\it Proceedings of the 24th International Joint Conference on Artificial Intelligence} (pp. 1305-1311). IJCAI.

[12] Yang, X., Khabsa, M., Wang, M., Wang, W., Awadallah, A., Kifer, D.\ \& Giles, C. L.\ (2018). Adversarial training for community question answer selection based on multi-scale matching. {\it arXiv preprint arXiv:1804.08058}.

[13] Wang, D.\ \& Nyberg, E.\ (2015). A long short-term memory model for answer sentence selection in question answering. In  {\it Proceedings of the 53rd Annual Meeting of the Association for Computational Linguistics and the 7th International Joint Conference on Natural Language Processing} (Vol. 2, pp. 707-712). ACL.

[14] Tan, M., Santos, C. D., Xiang, B.\ \& Zhou, B.\ (2015). Lstm-based deep learning models for non-factoid answer selection. {\it arXiv preprint arXiv:1511.04108}.

[15] Hao, Y., Liu, X., Wu, J.\ \& Lv, P.\ (2019). Exploiting Sentence Embedding for Medical Question Answering. In {\it Proceedings of the 33rd AAAI Conference on Artificial Intelligence}. AAAI.

[16] Wu, F., Duan, X., Xiao, J., Zhao, Z., Tang, S., Zhang, Y.\ \& Zhuang, Y.\ (2017). Temporal interaction and causal influence in community-based question answering. {\it IEEE Transactions on Knowledge and Data Engineering} 29(10), 2304-2317.

[17] Sha, L., Zhang, X., Qian, F., Chang, B.\ \& Sui, Z.\ (2018). A multi-view fusion neural network for answer selection. In {\it Proceedings of the 32nd AAAI Conference on Artificial Intelligence} (pp. 5422-5429). AAAI.

[18] Santos, C. D., Tan, M., Xiang, B.\ \& Zhou, B.\ (2016). Attentive pooling networks. {\it arXiv preprint arXiv:1602.03609}.

[19] Huang, H., Wei, X., Nie, L., Mao, X.\ \& Xu, X. S.\ (2018). From Question to Text: Question-Oriented Feature Attention for Answer Selection. {\it ACM Transactions on Information Systems (TOIS)} 37(1), 6.

[20] Zhao, Z., Lu, H., Zheng, V. W., Cai, D., He, X.\ \& Zhuang, Y.\ (2017). Community-based question answering via asymmetric multi-faceted ranking network learning. In {\it Proceedings of the 31st AAAI Conference on Artificial Intelligence} (pp. 3532-3538). AAAI.

[21] Wen, J., Ma, J., Feng, Y.\ \& Zhong, M.\ (2018). Hybrid Attentive Answer Selection in CQA With Deep Users Modelling. In {\it Proceedings of the 32nd AAAI Conference on Artificial Intelligence} (pp. 2556-2563). AAAI.

[22] Wen, J., Tu, H., Cheng, X., Xie, R.\ \& Yin, W.\ (2019). Joint modeling of users, questions and answers for answer selection in CQA. {\it Expert Systems with Applications} 118, 563-572.

[23] Zhang, X., Li, S., Sha, L.\ \& Wang, H.\ (2017). Attentive interactive neural networks for answer selection in community question answering. In {\it Proceedings of the 31st AAAI Conference on Artificial Intelligence} (pp. 3525-3531). AAAI.

[24] Zhao, Z., Zhang, L., He, X.\ \& Ng, W.\ (2014). Expert finding for question answering via graph regularized matrix completion. {\it IEEE Transactions on Knowledge and Data Engineering} 27(4), 993-1004.

[25] Hochreiter, S.\ \& Schmidhuber, J.\ (1997). Long short-term memory. {\it Neural computation} 9(8), 1735-1780.

[26] Graves, A.\ \& Schmidhuber, J.\ (2005). Framewise phoneme classification with bidirectional LSTM and other neural network architectures. {\it Neural Networks} 18(5-6), 602-610.

[27] Graves, A.\ (2013). Generating sequences with recurrent neural networks. {\it arXiv preprint arXiv:1308.0850}.

[28] Mikolov, T., Chen, K., Corrado, G.\ \& Dean, J.\ (2013). Efficient estimation of word representations in vector space. {\it arXiv preprint arXiv:1301.3781}.

[29] Hu, B., Lu, Z., Li, H.\ \& Chen, Q.\ (2014). Convolutional neural network architectures for matching natural language sentences. In {\it Advances in neural information processing systems} (pp. 2042-2050). NIPS.

[30] Tan, M., Dos Santos, C., Xiang, B.\ \& Zhou, B.\ (2016). Improved representation learning for question answer matching. In {\it Proceedings of the 54th Annual Meeting of the Association for Computational Linguistics} (Vol. 1, pp. 464-473). ACL.

[31] Yin, W., Yu, M., Xiang, B., Zhou, B.\ \& Schütze, H.\ (2016). Simple question answering by attentive convolutional neural network. {\it arXiv preprint arXiv:1606.03391}.

[32] Bian, W., Li, S., Yang, Z., Chen, G.\ \& Lin, Z.\ (2017). A compare-aggregate model with dynamic-clip attention for answer selection. In {\it Proceedings of the 2017 ACM on Conference on Information and Knowledge Management} (pp. 1987-1990). ACM.

[33] Xiang, Y., Chen, Q., Wang, X.\ \& Qin, Y.\ (2017). Answer selection in community question answering via attentive neural networks. {\it IEEE Signal Processing Letters} 24(4), 505-509.

[34] Resnick, P., Iacovou, N., Suchak, M., Bergstrom, P.\ \& Riedl, J.\ (1994). GroupLens: an open architecture for collaborative filtering of netnews. In {\it Proceedings of the 1994 ACM conference on Computer supported cooperative work} (pp. 175-186). ACM.

[35] Marlin, B. M.\ (2004). Modeling user rating profiles for collaborative filtering. In {\it Advances in neural information processing systems} (pp. 627-634). NIPS.

[36] Nakatsuji, M., Fujiwara, Y., Uchiyama, T.\ \& Toda, H.\ (2012). Collaborative filtering by analyzing dynamic user interests modeled by taxonomy. In {\it International Semantic Web Conference} (pp. 361-377). ISWC.

[37] Golub, G. H.\ \& Reinsch, C.\ (1971). Singular value decomposition and least squares solutions. In {\it Linear Algebra} (pp. 134-151). Berlin: Springer.

[38] Weston, J., Bengio, S.\ \& Usunier, N.\ (2011). Wsabie: Scaling up to large vocabulary image annotation. In {\it The 22nd International Joint Conference on Artificial Intelligence} (pp. 2764-2770). IJCAI.

[39] Deng, Y., Xie, Y., Li, Y., Yang, M., Du, N., Fan, W.\ \& Shen, Y.\ (2018). Multi-Task Learning with Multi-View Attention for Answer Selection and Knowledge Base Question Answering. {\it arXiv preprint arXiv:1812.02354}. 

[40] Nakov, P., Hoogeveen, D., Màrquez, L., Moschitti, A., Mubarak, H., Baldwin, T.\ \& Verspoor, K.\ (2017). SemEval-2017 task 3: Community question answering. In {\it Proceedings of the 11th International Workshop on Semantic Evaluation (SemEval-2017)} (pp. 27-48). ACL.

[41] Wei, X., Huang, H., Nie, L., Zhang, H., Mao, X. L.\ \& Chua, T. S.\ (2016). I know what you want to express: sentence element inference by incorporating external knowledge base. {\it IEEE Transactions on Knowledge and Data Engineering} 29(2), 344-358.

[42] Huang, X., Zhang, J., Li, D.\ \& Li, P.\ (2019). Knowledge graph embedding based question answering. In {\it Proceedings of the Twelfth ACM International Conference on Web Search and Data Mining} (pp. 105-113). ACM.

[43] Rotmensch, M., Halpern, Y., Tlimat, A., Horng, S. \ \& Sontag, D. A. \ (2017). Learning a health knowledge graph from electronic medical records. {\it Scientific Reports}, 7(1).

\end{document}